 \documentclass[preprint,3p,review,12pt]{elsarticle}

\usepackage{amssymb}
\usepackage{amsmath}

\usepackage{color}
\definecolor{redcolor}{rgb}{1.0,0.,0.}

\journal{}

\begin{document}

\begin{frontmatter}



\title{Word sense disambiguation via bipartite representation of complex networks}


\author{Edilson A. Correa Jr., Alneu de Andrade Lopes and Diego R. Amancio}

\address{Institute of Mathematics and Computer Science \\
University of S\~ao Paulo (USP)\\
S\~ao Carlos, S\~ao Paulo, Brazil}

\begin{abstract}
The word sense disambiguation (WSD) task aims at identifying the meaning of words in a given context for specific words conveying multiple meanings. This task plays a prominent role in a myriad of real world applications, such as machine translation, word processing and information retrieval. Recently, concepts and methods of complex networks have been employed to tackle this task by representing words as nodes, which are connected if they are semantically similar.
Despite the increasingly number of studies carried out with such models, most of them use networks just to represent the data, while the pattern recognition performed on the attribute space is performed using traditional learning techniques. In other words, the structural relationship between words have not been explicitly used in the pattern recognition process.
In addition, only a few investigations have probed the suitability of representations based on bipartite networks and graphs (bigraphs) for the problem, as many approaches consider all possible links between words.
In this context, we assess the relevance of a bipartite network model representing both feature words (i.e. the words characterizing the context) and target (ambiguous) words to solve ambiguities in written texts. Here, we focus on the semantical relationships between these two type of words, disregarding the relationships between feature words.
In special, the proposed method not only serves to represent texts as graphs, but also constructs a structure on which the discrimination of senses is accomplished.
Our results revealed that the proposed learning algorithm in such bipartite networks provides excellent results mostly when \emph{topical} features are employed to characterize the context. Surprisingly, our method even outperformed the support vector machine algorithm in particular cases, with the advantage of being robust even if a small training dataset is available. Taken together, the results obtained here show that the proposed representation/classification method might be useful to improve the semantical characterization of written texts.
\end{abstract}


\end{frontmatter}


\section{Introduction}
\label{}

The word sense disambiguation (WSD) task has been widely studied in the field of Natural Language Processing (NLP)~\cite{manningfoundations}. This task is defined as the ability to computationally detect which sense is being conveyed in a particular context~\cite{navigli2009word}. Although humans solve ambiguities in an effortlessly manner,
this matter remains an open problem in computer science, owing to the complexity associated with the representation of human knowledge in computer-based systems~\cite{mallery1988thinking}.
%
%
The importance of the WSD task stems from its essential role in a variety of real world applications, such as machine translation~\cite{weaver1955translation}, word processing \cite{church1995commercial}, information retrieval and extraction~\cite{stokoe2003word,markert2007semeval,zhou2005survey,FernandezAmoros20119506,Spina20134986,Fernández20129207}. In addition, the resolution of ambiguities plays a pivotal role in the development of the so-called semantic web~\cite{berners2001semantic}.

Many approaches devised to solve ambiguities in texts employ machine learning methods to automatically extract the best features in specific contexts~\cite{navigli2009word}. Automatic methods commonly use texts as a source of information, and these texts need to be transformed into a structured format. Popular representations are vectors of features, trees and graphs of relations between words~\cite{manningfoundations}. All such representations attempt to grasp, in a particular way, the semantical features related to the context surrounding ambiguous (target) words. Then, the information extracted from the context is used in the learning process.
Although graphs have been employed in general pattern recognition methods~\cite{5871621,Machicao201212626} and, particularly in the analysis of the semantical properties of texts in several ways~\cite{amancio2012unveiling,mihalcea2011graph,veronis2004hyperlex,voynich,7140830,Liu20083048,PhysRevE.74.026102}, the use of network models in the learning process has been restricted to a few works (see e.g.~\cite{0295-5075-98-5-58001,Silva:1556275}). In addition, most of the current network models emphasise the relationship between \emph{all} words of the document. As a consequence, a minor relevance has been given to the relationships between feature and target words. In this paper, we propose a different network representation which does not consider the relationship between all words, as described e.g. in~\cite{veronis2004hyperlex,0295-5075-98-5-58001}. We rather model texts using a bipartite network representation which focus on the relevant information arising from the relationship between \emph{feature} and \emph{target} words. This representation is then used as an underlying structure on which the proposed learning algorithm is applied. As we shall show, the combination
of this textual representation and the proposed learning technique may improve the classification process when compared with well-known supervised algorithms hinging on traditional text representations. Remarkably, we have also found that our method retains its discriminative power even when a considerable small amount of training instances is available.


%
%
%

The remainder of this paper is organized as follows. Section \ref{sec2} presents a brief review of basic concepts employed in this paper and related works. Section \ref{sec3} presents the details of the proposed representation and algorithm to undertake the word sense disambiguation task. In Section \ref{sec4}, we discuss the details of the experiments and the results concerning the accuracy and robustness of the proposed method. Finally, we present some perspectives for further works.

\section{Related works} \label{sec2}

The word sense disambiguation task can be defined as follows. Given a document represented as a sequence of words $T = \{w_1,w_2,...,w_n\}$, the objective is to assign appropriate sense(s) to all or some of the words $w_i \in T$. In other words, the objective is to find a mapping $A$ from words to senses, such that $A(w_i) \subseteq \mathcal{S}_D(w_i)$, where $\mathcal{S}_D(w_i)$ is the set of senses encoded in a dictionary $D$ for the word $w_i$, and $A(w_i)$ is the subset of appropriate senses of $w_i \in T$.
One of the most popular approaches to tackle the WSD problem is the use of machine learning, since this task can be seen as a supervised classification problem, where senses represent the classes~\cite{navigli2009word}. The attributes used in the learning methods are usually any informative evidence obtained from the topical context and external knowledge sources.
The latter approach is usually not common in practice because the creation of knowledge datasets demands  a time-consuming effort, since the change in domains requires  the recreation of new knowledge bases.


The generic WSD task can be distinguished into two types: \emph{lexical sample} and \textit{all-words} disambiguation. In the former, a WSD system is required to disambiguate a restricted set of target words. This is mostly done by supervised classifiers~\cite{navigli2009word}. In the \emph{all-words} scenario, the WSD system is expected to disambiguate all open-class words in a text. This task usually requires a wide-coverage of domains, and for this reason  a knowledge-based system is usually employed. In this article, only the \textit{lexical sample} task is considered.

The main step in any supervised WSD system is the representation of the context in which target words occur. The set of features employed typically are chosen to characterize the context in a myriad of forms~\cite{navigli2009word}. The most common types of attributes used for this aim are:
\begin{itemize}
\item \textit{local features}: the features of an ambiguous concept are a small number of words surrounding target words. The number of words representing the context is defined in terms of the window size $\omega$. For example, if the context of the target word $\tau_\omega$ is ``$p_{-3}$ $p_{-2}$ $p_{-1}$ $\tau_\omega$ $p_{+1}$ $p_{+2}$ $p_{+3}$'' and $\omega=2$, then the words $p_{-2}$, $p_{-1}$, $p_{+1}$ and $p_{+2}$ are used as features.
\item \textit{topical features}: the features are defined as topics of a text or discourse, usually denoted  in a bag-of-words representation;

\item \textit{syntatical features}: the features are syntactic cues and argument-head relations between the target word and other words within the same sentence; and
\item \textit{semantical features}: the features of a word are any semantic information available, such as previously established senses or domain indicators.
\end{itemize}
Using the aforementioned set of features, each word occurrence can be converted to a feature vector, which in turn is used as input in supervised classification algorithms. Typical classifiers employed for this task include decision trees~\cite{mooney1996comparative}, bayesian classifiers~\cite{mooney1996comparative,escudero2000portability}, neural networks~\cite{mooney1996comparative} and support vector machines~\cite{escudero2000portability,lee2002empirical}.

Another approach that has been used to address the WSD problem consists in the use of complex networks and graphs~\cite{mihalcea2011graph}. For instance, the HyperLex algorithm~\cite{veronis2004hyperlex} connects words co-occurring in paragraphs to establish similarity relations among words appearing in the same context. The frequency of co-occurrences is considered according to the following weighting scheme:
\begin{equation}
	w_{ij} = 1 - \max\{ P(w_i,w_j), P(w_j , w_i ) \}
\end{equation}
where $P(w_i,w_j) =  f_{ij}/f_i$, $f_i$ is the frequency of word $i$ in the document and $f_{ij}$
is the frequency of the co-occurrence of the words $i$ and $j$. Then, this network is used to create a tree-like structure via recognition of central concepts, which represent all possible senses. To perform the classification, the distance of context words to the central concepts in the tree structure is computed to identify the most likely sense.

Using a different approach, \cite{amancio2012unveiling} uses the local topological properties of co-occurrence networks to disambiguate target words. In this case, even though a significant performance has been found for particular target words, the optimal discrimination rate was obtained with traditional local features, suggesting thus that the overall discriminability could be improved upon combining features of distinct nature, as suggested by  similar approaches~\cite{WachsLopes20168,10.1371/journal.pone.0136076,0295-5075-100-5-58002}.

Despite the  numerous studies devoted to the WSD problem, this task remains an open problem in NLP, and currently it is considered one of the most complex problems in Artificial Intelligence~\cite{mallery1988thinking}. Our contribution in this paper is the proposition of a new representation that is able to focus the sense discrimination analysis on the relationship between features and target words. Unlike previous studies~\cite{amancio2012unveiling,veronis2004hyperlex}, we disregard the links between features words in our bipartite graph representation. Despite its seemingly simplicity, we show that such representation captures, in a artlessly manner, informative properties of target words and their respective senses. 

\section{Overview of the technique}\label{sec3}

This section presents the approaches to represent local and topic features of target words in a bipartite heterogeneous network. Here we also present the Inductive Model Based on Bipartite Heterogeneous Network (IMBHN) algorithm, which is responsible for inducing a classification model from the structure of a bipartite network~\cite{0295-5075-67-3-349,rossi2014inductive}.

\subsection{Modelling word context as a bipartite heterogeneous network}

Traditionally, the context of ambiguous words is represented in a vector space model, so that each target word is characterized by a vector. In this representation, each dimension of the vector corresponds to a specific feature. Alternatively, we may represent the data using a bipartite heterogeneous network. In this model, while the first layer comprises only feature words, the second only stores target words.
As mentioned in Section \ref{sec2}, currently, there exists a wide variety of features to tackle the WSD problem. In this paper, we focused on the analysis of \emph{local} and \emph{topical} attributes, as such data are readily available on (or derivable from) any corpus. Note that, in this case, we have not used any knowledge dataset.

In the proposed strategy based on \emph{topical} features, we create a set  $\mathcal{T}$ of   topical words. Then, each distinct becomes a distinct feature. As topical words, we considered the most frequent words of the dataset. The number of topical words, i.e. $|\mathcal{T}|$, is a free parameter. Given $\mathcal{T}$, the bipartite network is created by establishing a link between topical and target words whenever they co-occur in the {same document}.

In the proposed representation based on \emph{local} features, each feature word surrounding the target word represents an attribute. For each instance of the target word in the text, we select the $\omega$ closest surroundings words to become a feature word (see definition in Section \ref{sec2}). The selected words are then connected to the target words by weighted edges.

\subsection{Algorithm description}

The IMBHN algorithm can be used in the context of any text classification task. If the objective is to classify distinct documents in a given number of classes, the bipartite network can be constructed so that nodes represent both terms and documents. In this general scenario, such representation is used to compute the relevance of specific terms for distinct document classes. In a similar fashion, in this study, we compute the relevance of \emph{local}/\emph{topical} features for each target word. Then, this relevance is used to infer word senses.

\begin{figure}[ht]
	   \centering
       \includegraphics[width=0.85\textwidth]{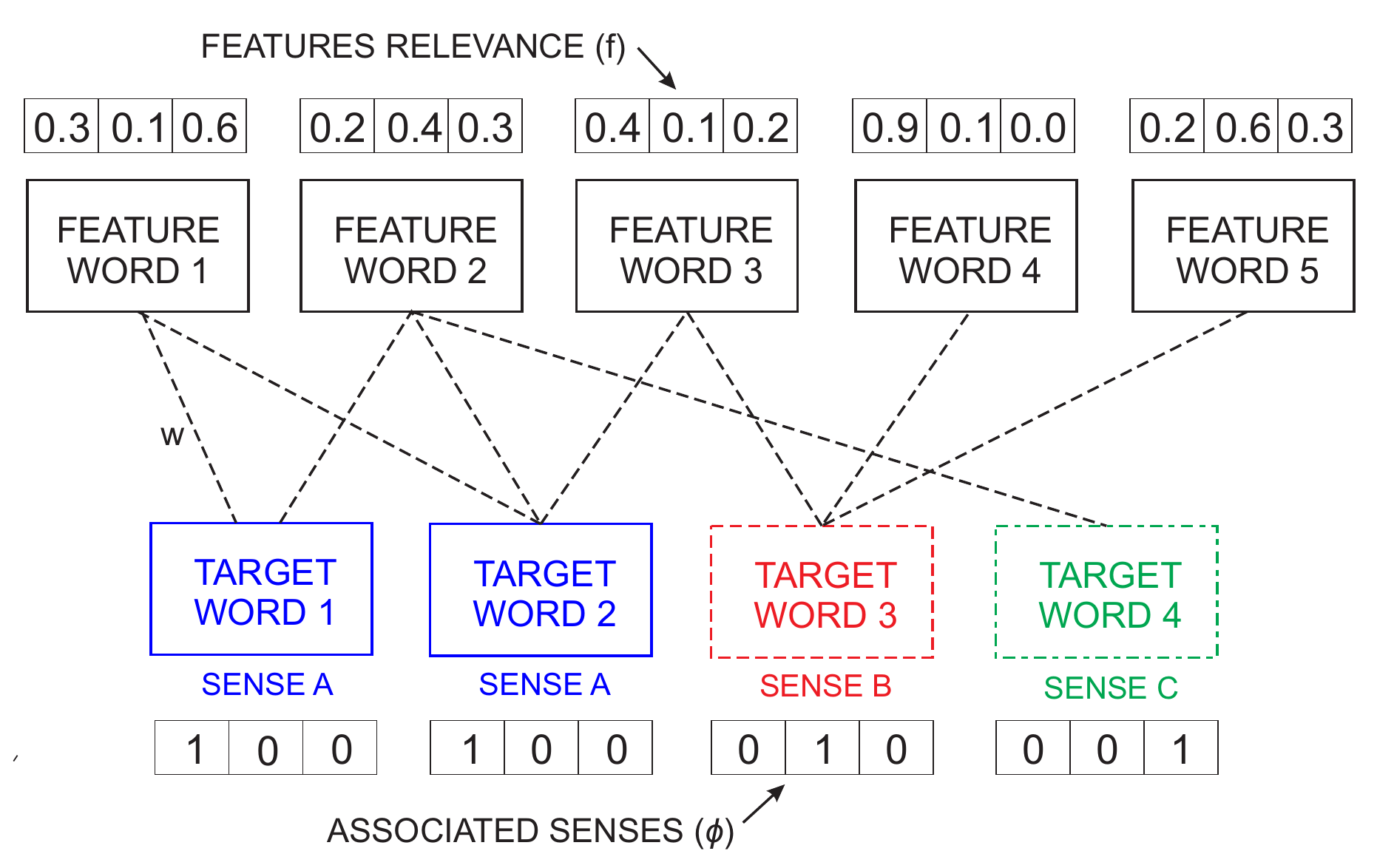}
       \caption{Bipartite network structure used by the IMBHN algorithm. Note the existence of two layers: the layer comprising feature words and the layer comprising target words, which can be classified into three distinct senses (A, B and C). For each feature word, there exists a vector of features relevance whose element $f_{t_i,c_j}$ denotes the relevance of $i$-th feature word for the $j$-th possible sense. The vectors below each target word represents the sense obtained in each iteration (i.e. $\phi_{d_k,c_j}$).}
       \label{fig1}
\end{figure}

The proposed algorithm for sense identification relies upon a network structure with two distinct layers: (i) a layer representing possible feature words (i.e. \emph{local} or \emph{topical} features), and (ii) a layer comprising all occurrences of the target word. The two layers are illustrated in Figure \ref{fig1}. Edges are established across layers so that context words and distinct occurrences of the target word are connected.  In addition, in the proposed network representation, a weight relating each feature word to each target word is also established. The main components of the model are:
\begin{itemize}
\item $w_{d_k,t_i}$:  the weight of the connection linking the $k$-th target word and the $i$-th feature word. In the strategy based on \emph{topical} features, this weight is constant along the execution of the algorithm and, for a given document $T$, is computed as
\begin{equation}\label{eq1}
   w_{d_k,t_i} = 1 - \delta(d_k,t_i) / l(T),
\end{equation}
where $\delta(d_k,t_i)$ denotes the the distance between two words (i.e. the number of intermediary words) and $l(T)$ is the length of $T$ (measured in terms of word counts). In the strategy based on \emph{local} features, the weight of the links is given by the term frequency - inverse document (tf-idf) strategy~\cite{manningfoundations}.

\item $f_{t_i, c_j}$: let $\mathcal{C}$ be the set of possible classes (i.e. word senses). $f_{t_i, c_j}$ represents the current relevance of the $i$-th feature word ($t_i \in \mathcal{T}$)  to the $j$-th class ($c_j \in \mathcal{C}$). This value is initialized using a heuristic and then is updated at each step of the algorithm. 

\item $y_{d_k,c_j}$: represents the \emph{actual} membership of the $k$-th target word. In other words, this is the label provided in the supervised classification scheme. If $c_j$ is the class of the $k$-th target word, then $y_{d_k,c_j} = 1$; otherwise, $y_{d_k,c_j} = 0$.

\item $\phi_{d_k,c_j}$: represents the \emph{obtained} membership of the $k$-th target word. If  $c_j$ is the class obtained for the $k$-th target word, then $\phi_{d_k,c_j} = 1$; otherwise, $\phi_{d_k,c_j} = 0$.

\item $\epsilon_{d_k,c_j}$: denotes the error of the current iteration. It is computed as:
\begin{equation} \label{errocomp}
	\epsilon_{d_k,c_j} = y_{d_k,c_j} - \phi_{d_k,c_j}.
\end{equation}
As we shall show, this error is used to update weights in $f$ so that, at each new iteration,
the distance between $y_{d_k,c_j}$ and $\phi_{d_k,c_j}$ decreases.
\end{itemize}
Note that, in the model illustrated in Figure \ref{fig1}, we only consider the relationship between feature and target words. Differently from traditional models, the relationship between feature words~\cite{amancio2012unveiling} is not explicitly considered in our model.

The training phase of the algorithm can be divided into the three following major steps:

\begin{enumerate}

\item {\bf Initialization}: there are three possible ways of initializing $f$, i.e. the vector weights of feature words.  The most simple strategy is to initialize weights with zeros or random values. A more informed alternative initializes weights using the a priori likelihood of feature words co-occur with senses. This probability can be computed as
\begin{equation}
	\textrm{Pr} = P(f_i|d_k) = {n_{f_i,d_k}}/{n_{d_k}},
\end{equation}	
where $n_{f_i,d_k}$ is the number of times that the $i$-th feature word appears in the context of the $k$-th target word and $n_{d_k}$ is the total number of occurrences of $d_k$. In our experiments,  we report  the best results obtained among these three alternatives.
\ \\

\item {\bf Error calculation}:
In the error calculation step, firstly, the output vector for each target word ($\phi(d_k)$) is computed.
This vector depends upon the presence of the feature word in the context ($w_{d_k, t_i}$) and its relevance for the class ($f_{t_i, c_j}$). Mathematically, the class computed at each new iteration is given by
%
%
%
\begin{equation} \label{eq:clas}
 C\Bigg{(} \sum_{t_i \in \mathcal{T}} w_{d_k, t_i}  f_{t_i, c_j}\Bigg{)} = \\
 \begin{cases}
	 	 1, & \textrm{if } c_j = \arg\max \limits_{c_l \in \mathcal{C} } \Big{(} \sum\limits_{t_i \in \mathcal{T} } w_{d_k, t_i}  f_{t_i, c_l} \Big{)}. \\
 	        0, & \textrm{otherwise.}
 \end{cases}
\end{equation}
%
%
After updating the classes for each target word, the values of $f_{t_i, c_j}$ are modified. This update is controlled by the correction rate $\eta$:
\begin{equation} \label{atualiza}
f_{t_i,c_j}^{(n+1)} = f_{t_i,c_j}^{(n)} + \eta  \sum_{d_k \in \mathcal{D}} w_{d_k,t_i}  \epsilon_{d_k,c_j} ^{(n)},
\end{equation}
where the superscipt $(n)$ in $f$ and $\epsilon$ denotes the value of these quantities computed in the $n$-th iteration of the algorithm and $\mathcal{D}$ is the set of target words. Note that $\epsilon_{d_k,c_j} ^{(n)}$ is computed as defined in equation \ref{errocomp}.
The values of $\epsilon_{d_k,c_j}$ and $f_{t_i,c_j}$ in equations \ref{errocomp} and \ref{atualiza}, respectively, are updated until a stop criterion is reached. In our experiments, we have stopped the algorithm when a minimum error $\epsilon_{min} = 0.01$ is obtained. If the minimum error is not reached after $n_{max} = 1,000$ iterations, the algorithm is stopped.


%
%

%

\ \\

\item {\bf Classification}: in the classification phase, the induced values of $f$ are used in the classification. The  word senses for each ambiguous word of the dataset are then obtained by computing the following linear combination:
\begin{equation}
\textrm{class}(d_k) = \arg\max \limits_{c_j \in \mathcal{C} } \Bigg{(} \sum\limits_{t_i \in \mathcal{T} } w_{d_k, t_i}  f_{t_i, c_j} \Bigg{)}.
\end{equation}

\end{enumerate}

\section{Experimental Evaluation}\label{sec4}
This section presents the corpus used in the experiments. In addition, we also detail the experimental configuration of parameters. Finally, we present a robustness analysis to investigate how the performance of the IMBHN varies with the size of the training set.


\subsection{Corpus}

In order to evaluate the proposed algorithm, the SENSEVAL-2~\cite{edmonds2001senseval} corpus was used. This corpus comprises documents from distinct sources, including the British National Corpus and the Penntreebank portion of the Wall Street Journal. The SENSEVAL-2 corpus encompasses $15,225$ instances of short texts representing the context surrounding  ambiguous words. Each word is tagged with its part-of-speech, and the manually annotated senses of four target words is provided. The number of senses and the number of instances of each word used in our experiments is shown in Table \ref{tabsenseval}. In the evaluation process, these four words were considered as the target words. 
In particular, to characterized the contexts, we have removed stopwords and punctuation marks as such elements do not convey any semantical meaning and, therefore, do not improve the characterization of contexts.
%
\begin{table}[h]
\centering
\caption{List of words used to evaluate the proposed word sense disambiguation algorithm. NS and NI denote the number of senses of the target word and the number of instances in the corpus, respectively. The dataset comprising word context and word senses was obtained from the SENSEVAL-2 corpus~\cite{edmonds2001senseval}. Prior to the application of the learning methods, stopwords and punctuation marks were removed from the original instances.}
\begin{tabular}{|l|c|c|c|}
\hline
{\bf Target word} & {\bf NS} & {\bf NI} \\
\hline
interest (noun) & 6 & 2,368 \\
line (noun) & 6 & 4,146 \\
serve (verb) & 4 & 4,378 \\
hard (adjective) & 3 & 4,333 \\
\hline
\end{tabular}
\label{tabsenseval}
\end{table}
%

\subsection{Experiment Configuration}

The results obtained by the IMBHN algorithm were compared with four inductive classification algorithms: Naive Bayes (NB)~\cite{Caruana:2006:ECS:1143844.1143865}, J48 (C4.5 algorithm)~\cite{Quinlan1993}, IB\textit{k} (\textit{k}-Nearest Neighbors)~\cite{Aha:1991:ILA:104713.104717} and Support Vector Machine via sequential minimal optimization  (SMO)~\cite{Platt1998}. The parameters of these algorithms have been chosen using the methodology described in~\cite{amancio2014systematic}.  For the IMBHN algorithm, we used the error correction rates $\eta=\{0.01, 0.05,0.10,0.50\}$. The number of topical features used in the experiments were $|\mathcal{T}|=\{100, 200, 300\}$. Finally,  the window size for the local features were $\omega=\{1,2\}$. The evaluation process was performed via 10-fold cross-validation~\cite{Kohavi:1995:SCB:1643031.1643047}.

\subsection{Results and discussion}

To analyze the behavior and accuracy of the proposed algorithm, we first studied the WSD task using topical features to characterize the context of target words of our dataset. The obtained results are shown in Table \ref{tab2}. When the number of topical features $|\mathcal{T}|$ is set with $|\mathcal{T}|=100$, the best results occurred for the SMO and J48 techniques. In three cases, the proposed algorithm IMBHN performed worse than the best results achieved with competing techniques.
\begin{table}
\centering
\caption{\label{tab2}Accuracy rates obtained by each algorithm using \emph{topical} features to disambiguate the following target words: (i) ``interest'' (noun), (ii) ``line'' (noun), (iii) ``serve'' (verb) and (iv) ``hard''. The best results for each value of $|\mathcal{T}|$ and for each target word are highlighted in bold font. The best results tend to occur with the SMO method, however, in particular cases, the J48 outperforms the SMO learning technique. Apart from the word ``serve'' when $|\mathcal{T}|=300$, the IMBHN does not perform as good as the other traditional methods. }
\begin{tabular}{|l|c|c|c|c|c|}
\hline
{\bf Method} & $|\mathcal{T}|$ & interest & line & serve & hard\\
\hline
 IMBHN 	& 100 & 71.49\% & 59.91\% & 64.68\% & 77.28\% \\
 J48 	& 100 & 79.47\% & 62.73\% & \textbf{68.15\%} & \textbf{84.58\%} \\
 IBk 	& 100 &  75.71\% & 53.18\% & 63.68\% & 79.34\% \\
 NB 		&100 & 59.79\% & 51.95\% & 58.79\% & 43.04\% \\
 SMO 	& 100 & \textbf{79.77\%} & \textbf{62.87\%} & 66.79\% & 84.07\% \\
\hline
IMBHN & 200 & 78.50\% & 65.53\% & 66.56\% & 78.74\% \\
J48 & 200 & 82.39\% & 66.71\% & 68.95\% & \textbf{86.17\%} \\
IBk & 200 & 80.70\% & 53.93\% & 63.24\% & 80.10\% \\
NB & 200 & 60.17\% & 54.43\% & 61.71\% & 42.69\% \\
SMO & 200 & \textbf{83.27\%} & \textbf{68.95\%} & \textbf{69.84\%} & 85.36\% \\
\hline
IMBHN 	& 300 & 80.23\% & 67.82\% & 71.42\% & 78.62\% \\
J48       	& 300 & 82.68\% & 68.54\% & 70.67\% & \textbf{86.22\%} \\
IBk	& 300 & 80.32\% & 54.05\% & 63.13\% & 80.38\% \\
NB 		& 300 & 55.66\% & 54.14\% & 66.99\% & 41.61\% \\
SMO      & 300 & \textbf{84.71\%} & \textbf{69.87\%} & \textbf{71.92\%} & 85.52\% \\
\hline
Baseline & -- & 52.80\% & 53.40\% & 41.40\% & 79.30\% \\
\hline
\end{tabular}
\end{table}

In general, the performance of the classifiers tend to improve when the number of topical features ($|\mathcal{T}|$) increases from $100$ to $300$. This is clear when one observes that e.g. the best accuracy rate for the word ``interest'' goes from  $79.77\%$ to $84.71\%$. The same behavior can be observed for the other target words of the dataset, however, in a minor proportion. Concerning the performance of the proposed technique when $|\mathcal{T}|=\{200,300\}$, in most cases, the IMBHN method is outperformed by the SMO technique, which provided the best results for the words ``interest'', ``line'' and ``serve''. The best results for the word ``hard'' was achieved with the J48 classifier.

When analyzing the performance of the classifiers induced with local features, a different pattern of  accuracy has been found, as shown in Table \ref{tab3}. For the words ``interest'', ``line''and ``serve'' the IMBHN classifier yielded the best results, for $\omega=\{1,2,3\}$. Conversely, if we consider the word ``hard'', the decision tree based algorithm, J48, outperformed all other methods. However, the performance achieved with J48 was very similar to the one obtained with the IMBHN: the maximum difference of accuracy between these two classifiers was $1.09\%$, when $\omega=3$. This observation confirms the suitability of the proposed method for the problem, as optimized results have been found for virtually all words of the dataset.


%
\begin{table}
\centering
\caption{Accuracy rates obtained by each algorithm using \emph{local} features to disambiguate the following target words: (i) ``interest'' (noun), (ii) ``line'' (noun), (iii) ``serve'' (verb) and (iv) ``hard''. The best results for each value of $\omega$ and for each target word are highlighted in bold font. For the words ``interest'', ``line'' and ``serve'', the best performance is achieved with the IMBHN method in all of the studied scenarios. For the word ``hard'', the J48 learning algorithm displayed the best performance. However, in this case, the IMBHN method performed almost as well as the J48, for $\omega =\{1,2,3\}$. Another interesting pattern arising from the results is the fact that  performances are improved when $\omega$ takes higher values.
}
\begin{tabular}{|l|c|c|c|c|c|}
\hline
{\bf Method} & $\omega$ & interest & line & serve & hard\\
 \hline
IMBHN 	& 1 & \textbf{81.50\%} & \textbf{69.19\%} & \textbf{69.96\%} & 85.50\% \\
J48 	& 1 & 65.83\% & 60.97\% & 46.43\% & \textbf{85.57\%} \\
IBk 	& 1 & 74.73\% & 59.76\% & 62.54\% & 82.06\% \\
NB 		& 1 & 64.90\% & 37.16\% & 42.11\% & 43.94\% \\
SMO 	& 1 & 66.00\% & 62.61 & 57.88\% & 81.30\% \\
\hline
IMBHN	& 2 & \textbf{83.27\%} & \textbf{75.80\%} & \textbf{78.48\%} & 84.67\% \\
J48 	& 2 &71.74\% & 61.21\% & 55.57\% & \textbf{85.39\%} \\
IBk	& 2 &65.32\% & 56.72\% & 58.26\% & 78.35\% \\
NB 		& 2 &66.97\% & 45.22\% & 60.16\% & 43.68\% \\
SMO 	& 2 & 64.10\% & 62.13 & 58.63\% & 80.68\% \\
\hline
IMBHN	& 3 & \textbf{85.55\%} & \textbf{77.13\%} & \textbf{80.12\%} & 84.16\% \\
J48 	& 3 & 76.85\% & 62.66\% & 60.94\% & \textbf{85.25\%} \\
IBk 	& 3 & 52.44\% & 53.59\% & 52.12\% & 78.86\% \\
NB 		& 3 & 68.49\% & 50.43\% & 66.05\% & 42.97\% \\
SMO 	& 3 & 64.14\% & 60.80 & 58.45\% & 79.78\% \\
\hline
Baseline & -- & 52.80\% & 53.40\% & 41.40\% & 79.30\% \\
\hline
\end{tabular}
\label{tab3}
\end{table}

The best results obtained with topical and local features are summarized in Table \ref{tab4}. The proposed algorithm for representing texts and discriminating senses outperformed other methods when considering also distinct types of features. In special, the IMBHN performed significantly better than the SMO method for the word ``line'' and ``serve''. A minor gain in performance has been observed for ``interest''. With regard to the word ``hard'', the best performance was obtained with the J48 (with topical features). However, a similar accuracy was obtained with the IMBHN (with local features, as shown in Table \ref{tab3}).  All in all, these results show, as a proof of principle, that the proposed algorithm may be useful to the word sense disambiguation problem, as optimal or near-optimal performance has been found in the studied corpus.

\begin{table}
\centering
\caption{Best classifiers for each feature set and its accuracy.}
\begin{tabular}{|l|c|c|}
\hline
{\bf Target word} & {\bf Topical features}  & {\bf Local features}  \\
\hline
interest (noun) 	& 84.71\% (SMO) & 85.55\% (IMBHN) \\
line (noun) 		& 69.87\% (SMO) & 77.13\% (IMBHN) \\
serve (verb) 		& 71.92\% (SMO) & 80.12\% (IMBHN)  \\
hard (adjective) 	& 86.22\% (J48)   & 85.57\% (J48) \\
\hline
\end{tabular}
\label{tab4}
\end{table}

\begin{figure}
	   \centering
       \includegraphics[width=1\textwidth]{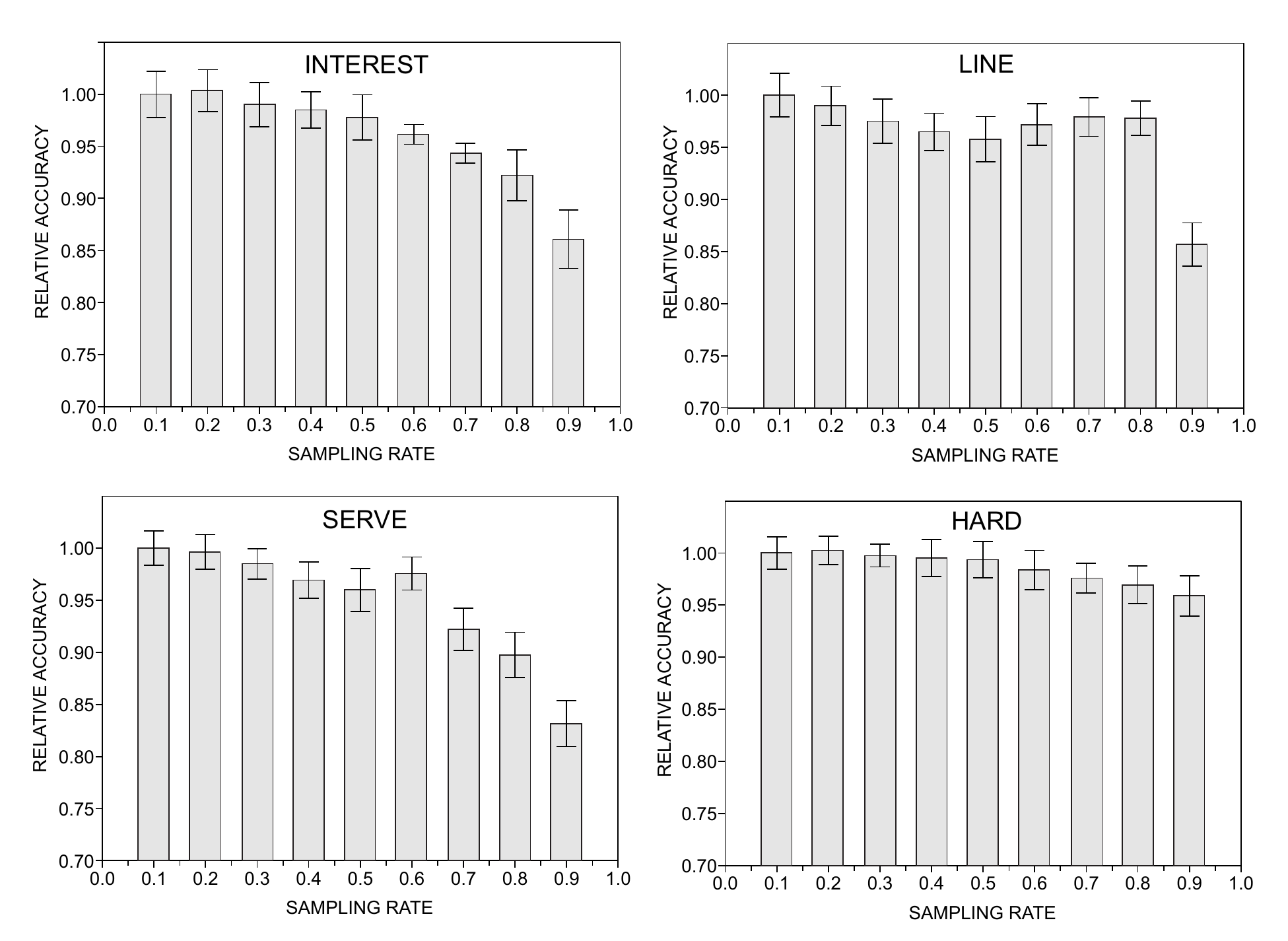}
       \caption{Robustness analysis performed with the IMBHN algorithm. The sampling rate corresponds to the fraction (percentage) of instances randomly removed from the original dataset. The relative accuracy is given by equation \ref{relativo}. Note that, in the worst case, the accuracy of the IMBHN reaches 85\% of the accuracy when only 10\% of the original data is available ($S=0.9$), confirming thus the robustness of the method. A similar behavior was obtained when the approach based on topical features was evaluated with $\omega=\{1,2\}$.}
       \label{figrob}
\end{figure}

\begin{figure}
	   \centering
       \includegraphics[width=1\textwidth]{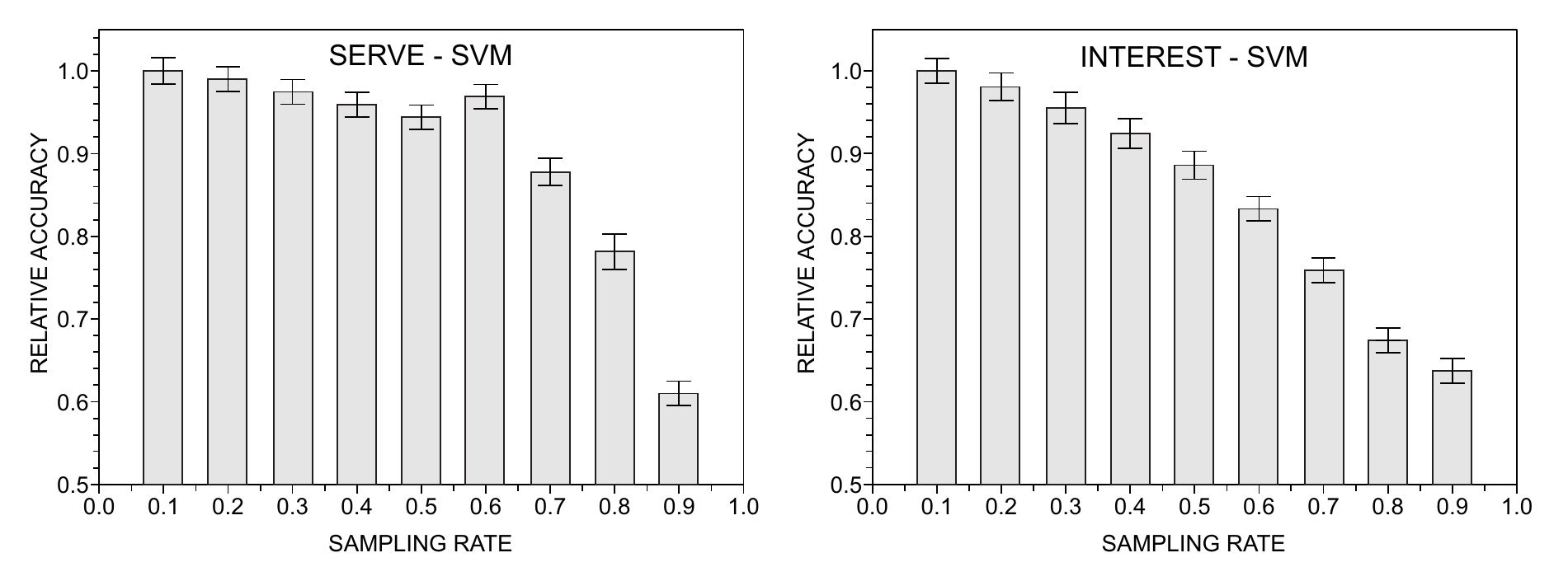}
       \caption{Robustness analysis performed with the SMO algorithm for two words of the dataset. The sampling rate corresponds to the fraction (percentage) of instances randomly removed from the original dataset. The relative accuracy is given by equation \ref{relativo}. Unlike the IMBHN algorithm, the accuracy rate drops significantly for high sampling rates. }
       \label{figrobsvm}
\end{figure}

A disadvantage associated to the use of supervised methods to undertake the word sense disambiguation problem is the painstaking, time-consuming effort required to build reliable datasets~\cite{navigli2009word}. For this reason, it becomes relevant to analyze the performance of WSD systems when only a few labelled instances are available for training~\cite{navigli2009word}. In this sense, we performed a robustness analysis of the proposed algorithm to investigate how performance is   affected when smaller fractions of the dataset are provided for the algorithm. To perform such a robustness analysis the following procedure was adopted. We defined a sampling rate $\mathcal{S}$, representing the percentage of \emph{disregarded} instances from the original dataset. For each sampling rate, we computed the accuracy $\Gamma(S)$ relative to the sampled dataset.  The relative accuracy rate for a given $S$ was computed as
\begin{equation} \label{relativo}
	\tilde{\Gamma}(S) = \frac{\Gamma(S)}{\Gamma(0)},
\end{equation}
which quantities the percentage of the original accuracy which is preserved when the original dataset is sampled with sampling rate $S$. For each sampling rate, we generated $50$ sampled subsets.  The obtained results for the IMBHN in its best configuration (i.e. using local features and $\omega=3$) are shown in
Figure \ref{figrob}. The best scenario occurs for the word ``hard'', as even when 90\% of the original is ignored, in average, more than 95\% of the original accuracy (i.e. $\Gamma(S=0)$) is recovered. Concerning the other words, a good performance was also observed when only a small fraction was available. This is the case of ``serve'': when 90\% of the dataset is disregarded, 85\% of the original accuracy is kept. These results suggest that the IMBHN could be successfully applied in much smaller datasets without a significative loss in performance. We have found similar robustness results for other configurations of parameters ($\omega$) of the IMBHN (results not shown), which reinforces the hypothesis that the resiliency of the method with regard to the total amount of instances in the training phase is stable with varying parameter values.
Note that such a robustness, although strongly desired in practical problems, does not naturally arise in all pattern recognition methods. This is evident e.g. when the robustness SMO is verified for ``serve'' and ``interest'', as shown in Figure \ref{figrobsvm}. Note that when $S=0.9$, the accuracy drops to about 60\% of its original value. 
%

\section{Conclusion}\label{sec5}

The accurate discrimination of word senses plays a pivotal role in information extraction and document classification tasks. While methods based on deep paradigms may perform well in 
very specific domains, statistical methods based mainly on machine learning have proved useful to undertake the word sense disambiguation task in more general contexts. In this article, we have devised a statistical model to both represent contexts and recognize patterns in written texts. The model hinges on a bipartite network, with layers representing feature words and target words, i.e. words conveying two or more potential senses. We have shown, as a proof of principle, that the proposed model presents a significant performance, mainly when contextual features are modelled via extraction of local words to represent semantical contexts. We have also observed that, in general, our method performs well even if a relatively small amount of data is available for the training process. This is an important property as it may significantly reduce both time and  effort required to construct a corpus of labelled data.

As future works, we intend to explore further generalizations of the algorithm. Owing to the power of word adjacency networks in extracting relevant semantical features of texts~\cite{amancio2012unveiling}, we intend to use such models to improve the characterization of the studied bipartite networks.  The word adjacency model could be used, for example, to better represent the relationship between feature and target words by using network similarity measurements~\cite{scisim,newsurvey,PhysRevE.73.026120}. We also intend to extend the present model to consider topological and dynamical measurements of word adjacency networks as local features~\cite{amancio2012unveiling}. 

\newpage

\section*{Acknowledgements}
 \noindent
E.A.C. Jr. and D.R.A. acknowledge financial support from Google (Google Research Awards in Latin America grant). D.R.A. thanks S\~ao Paulo Research Foundation (FAPESP) for support (grant. no. 2014/20830-0). A.A.L. acknowledges support from FAPESP (grant no. 2011/22749-8 and 2015/14228-9) and CNPq (Brazil) (grant no. 302645/2015-2).

%



\newpage

\bibliographystyle{model1-num-names}
\bibliography{annot}







\end{document}